\documentclass[conference]{IEEEtran}
\IEEEoverridecommandlockouts

\usepackage[utf8]{inputenc} 
\usepackage[T1]{fontenc}    
\usepackage[colorlinks]{hyperref}       
\usepackage{url}            
\usepackage{booktabs}       
\usepackage{amsfonts}       
\usepackage{nicefrac}       
\usepackage{microtype}      
\usepackage{lipsum}		
\usepackage{graphicx}
\usepackage{doi}
\usepackage{caption}
\usepackage{subcaption}
\usepackage{rotating}
\usepackage{algorithm}
\usepackage{algpseudocode}
\usepackage{bm} 
\usepackage{csquotes}
\usepackage{xcolor}
\usepackage{multirow}
\usepackage{hhline}
\usepackage{makecell}
\usepackage{colortbl}
\usepackage{tabu}
\usepackage{tabularx}
\usepackage{array}
\usepackage{adjustbox}
\usepackage{caption}
\usepackage[para]{footmisc}

\usepackage{float}
\usepackage{amsmath}

\algnewcommand{\LineComment}[1]{\State \(\triangleright\) #1} 

\begin{document}

\title{\enquote{Just Drive}: Colour Bias Mitigation for Semantic Segmentation in the Context of Urban Driving}

\author{\IEEEauthorblockN{Jack Stelling}
\IEEEauthorblockA{\textit{School of Computing}\\
Newcastle University, UK\\
jackstelling2@gmail.com}

\and
\IEEEauthorblockN{Amir Atapour-Abarghouei}
\IEEEauthorblockA{\textit{Department of Computer Science}\\
Durham University, UK\\
amir.atapour-abarghouei@durham.ac.uk}
}


\maketitle

\begin{abstract}

	Biases can filter into AI technology without our knowledge. Oftentimes, seminal deep learning networks champion increased accuracy above all else. In this paper, we attempt to alleviate biases encountered by semantic segmentation models in urban driving scenes, via an iteratively trained \emph{unlearning} algorithm. Convolutional neural networks have been shown to rely on colour and texture rather than geometry. This raises issues when safety-critical applications, such as self-driving cars, encounter images with covariate shift at test time - induced by variations such as lighting changes or seasonality. Conceptual proof of bias \emph{unlearning} has been shown on simple datasets such as MNIST. However, the strategy has never been applied to the safety-critical domain of pixel-wise semantic segmentation of highly variable training data - such as urban scenes. Trained models for both the baseline and bias unlearning scheme have been tested for performance on colour-manipulated validation sets showing a disparity of up to 85.50\% in mIoU from the original RGB images - confirming segmentation networks strongly depend on the colour information in the training data to make their classification. The bias unlearning scheme shows improvements of handling this covariate shift of up to 61\% in the best observed case - and performs consistently better at classifying the \enquote{human} and \enquote{vehicle} classes compared to the baseline model.

\end{abstract}
\vspace{3mm}


\section{Introduction}
\label{sec:intro}

Recent years have seen a surge in the development of artificial intelligence (AI) systems; largely fuelled by the symbiosis of deep learning progress \cite{girshick2013, atapour2018real, devlin2018bert, atapour2019king, badrinarayanan2017, atapour2020rank}, and advancements in micro/nanochip manufacturing –- harnessing remarkable compute power. Ubiquitous deployment of AI throughout modern society means that practitioners have an ethical and moral responsibility in the dissemination of this cutting-edge technology.

Within the field of deep learning, convolutional neural networks (CNNs) have gained substantial traction in the last decade, and their performance in computer vision tasks is state of the art \cite{chen2018, tan2020, kumar2020}. Due to the complex nature of these systems, a \enquote{black box} stigma is often attached to them; since deep learning networks are now achieving unprecedented levels of accuracy, external scrutiny and media spotlight demands a push towards transparency, fairness, and accountability \cite{lundberg2017, fong2017}. This has led to the more recent movement of \enquote{explainable artificial intelligence} (XAI).

The crux of developing fair AI is the elimination of bias -- a long sought issue in any statistical modelling application. Bias can manifest itself in a multitude of guises which are generally quite context specific. For the purposes of this work, we will focus on algorithmic bias. Bias of this nature can creep into our models from training data, meaning our models exhibit the same systemic discrimination found in the wild. This bias can often be unknown and undetected – making for a particularly insidious force.  Indeed, George Santayana warned us that \emph{\enquote{those who do not remember the past are condemned to repeat it}.} If bias \emph{is} allowed to creep into our models undetected, we risk propagating this bias throughout society; eventually distilling into a self-fulfilling prophecy –- one which automates inequality.  

\begin{figure}[t!]
	\centering
	\includegraphics[width=0.99\linewidth]{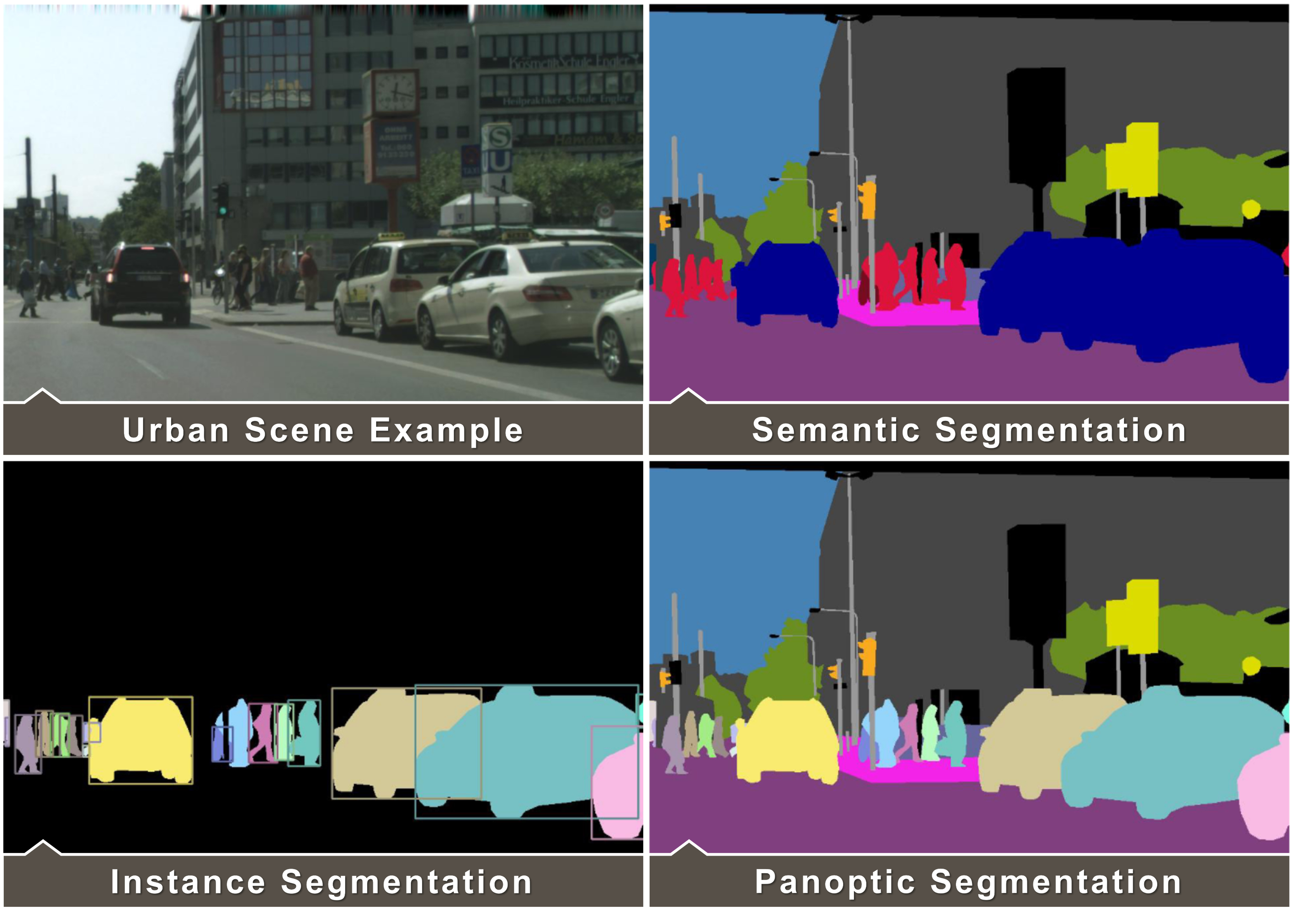}
	\captionsetup[figure]{skip=7pt}
	\captionof{figure}{Semantic, Instance and Panoptic segmentation \cite{kirillov2019}.}
	\label{fig:first}\vspace{-0.25cm}
\end{figure}

This research focuses on colour bias which exists within urban scenes. Urban scene segmentation is at the heart of autonomous vehicle (AV) technology \cite{hawke2021}, and to date, many network architectures have been developed achieving impressive accuracy \cite{badrinarayanan2017, chen2017b}. This blinkered push towards optimal accuracy often neglects to consider biases within the training data; thus, any advancements in the field of bias mitigation in particular within the realm of image segmentation are significant.

Colour bias can manifest in many ways within urban scenes. Successful AV technology must be dependable in a multitude of conditions, including densely populated urban streets, fog, rain, snow, glare, and seasonal changes. This highly variable distribution of data poses a great challenge. Clearly, the same tree that a segmentation model picks up on in summer might look very different in winter, or in New York would a CNN learn to categorise all yellow boxes as cars due to the number of taxis? We humans are extremely adaptable and can make decisions to correct our actions depending on external stimulus. Machines, however, struggle to perform well in edge cases or situations not encountered in the training phase. Consequently, a push towards robustness and generalisation is paramount for the evolution of safe AV technology.

The umbrella of image segmentation covers three main domains: instance segmentation \cite{he2017}, semantic segmentation \cite{chen2017b, badrinarayanan2017}, and more recently a hybrid of both – panoptic segmentation \cite{cheng2020} (Figure \ref{fig:first}).  In this work, we will consider semantic segmentation, which concerns categorising each pixel in an image into one of $n$ predefined classes. This process splits the image into different regions based on what the pixels show and is the adopted methodology for AV systems. While this work focuses on semantic segmentation, the same principles can be applied to instance and panoptic segmentation or in fact many other computer vision tasks \cite{atapour2018extended, papageorgiou1998general, torralba2002depth, atapour2018comparative}.

The aim of this paper is to mitigate colour bias from semantic segmentation models trained on urban driving scenes. The primary contributions of this work are twofold:

\begin{itemize}

\item We provide empirical evidence that seminal semantic segmentation architectures do overfit to the colour information in highly variable urban scenes and, where possible, attempt to quantify this. 

\item We also demonstrate that a multi-headed network architecture can adversarially remove a \emph{known} bias during training in a pixel-wise semantic segmentation model. 

\end{itemize}

In the interest of reproducibility, transparency and progression, all code is publicly available in a project repository\footnote{https://github.com/JackStelling/BiasMitigation}. 


\section{Related Work}
\label{section:related-work}

We consider related work within the context of Semantic Segmentation (Section \ref{subsection:seg-architectures}) and Bias Removal (Section \ref{subsection:bias-removal}). 


\subsection{Semantic Segmentation}
\label{subsection:seg-architectures} 

For the purposes of semantic segmentation, fully convolutional networks (FCN) have gained the most traction in recent years after the work of Long et al. \cite{long}. FCNs do not have any fully-connected layers, solely relying on convolutional layers passed to a classification layer. This preserves spatial information from the input and significantly reduces the number of parameters in the network, thus increasing computational efficiency.  

Encoder-decoder style networks achieve impressive results with U-Net \cite{ronneberger2015}, SegNet \cite{badrinarayanan2017} and DeConvNet \cite{hyeonwoo2015} all adopting the same idea, whereby the first half of the network is a mirror image of the second half, albeit with different methodical nuances. SegNet \cite{badrinarayanan2017} proposes a computationally fast and memory efficient network by saving the indices of the maximum values on the max-pooling operation – this is then used during upsampling in the decoder part of the network. U-Net utilises skip connections, allowing for information from the encoder part of the network to be used in the decoding procedure.   

Later, the DeepLab family \cite{chen2016, chen2017a, chen2017b, chen2018} of architectures pushed the boundaries in semantic segmentation using atrous convolutions \cite{chen2017a}, effectively developing the Atrous Spatial Pooling Pyramid (ASPP) module to handle objects of different scales in the same image. ASPP is based on simultaneously performing dilated convolutions with different atrous rates and concatenating the resultant feature maps. This essentially combines multiple fields of view, tackling the issue of different scale objects. The DeepLab family have achieved state-of-the-art results on benchmark datasets with each iteration of the network.  

More recently, the concept of attention has been applied to semantic segmentation tasks \cite{tao2020}, creating a more efficient and higher performing model than using multiscale inference. Similarly to the problem that DeepLab attempted to solve with the ASPP module, Tao et al. \cite{tao2020} argue that fine detail (bollards, a person in the distance, etc.) is often better predicted with a scaled-up image size. Whereas large objects (roads, buildings, etc.) require more global context and downscaled images are generally more beneficial as the convolutional filters have a larger field of view and thus capture more context. Tao et al. develop a system whereby prediction for some pixels is performed using the scaled-up images and others use the scaled down images. Further, the ASPP module and other multiscale context methods e.g., PSPNet \cite{zhao2017} are static and not learned, whereas relational methods build context based on image composition. This means that unlike \cite{chen2017a, chen2017b, chen2018, zhao2017}, the region of interest using an attention-based mechanism is not restricted to being square – this is advantageous in the context of urban scenes when the geometry is often a product of visual perspective, like a road sign in the foreground covering a long skinny rectangular patch. Tao et al. maintain state-of-the-art performance on the CityScapes \cite{cordts2016} and Mallipary Vistas \cite{neuhold2017} datasets at the time of writing.

In this work, we adopt the DeepLabV3 \cite{chen2017b} and SegNet \cite{badrinarayanan2017} architectures to use as baselines for testing novel bias removal concepts in the context of semantic segmentation. DeepLab architectures have been shown \cite{arnaba2018} to be less susceptible to adversarial attacks, an increasing concern in safety-critical applications such as self-driving cars. Other segmentation models, however, can similarly be used given the overall pipeline of our approach. 

\subsection{Bias Removal}
\label{subsection:bias-removal} 
\vspace*{-0.5mm}

As mentioned in Section \ref{sec:intro}, \emph{\enquote{bias}}, in the context of urban scene segmentation, can manifest in many forms. Examples include adverse weather conditions experienced at test time when training data does not account for this, seasonality affecting physical colours (e.g., tree leaves, flora), seasonality affecting lighting/shadows/luminance, more obvious lighting fluctuations from night to day, reflection, shadow and different countries/localities using different colour systems for highway codes, among others. This is by no means an exhaustive list, as even edge cases such as sporting events, parades and accident blockades can cause out-of-sample differences that networks must tackle if we are to trust AV technology. These unknown perturbations cause a covariate shift from the input data that the models are trained on, which can cause adverse effects on performance due to the high intercorrelation between network weights.      

Large-scale, finely annotated datasets for segmentation are expensive to obtain, requiring human annotation which can often take experienced workers up to 90 minutes an image to complete \cite{cordts2016}. Due to this bottleneck, it is not feasible to create site specific training datasets for multiple locations where the cars will operate, thus robust generalisable models must be developed which perform well in a wide variety of situations. Models which can learn bias in the data and account for it are highly desirable. Not only does this problem exist within the sphere of AV technology, it also extends to many others, including the fields of augmented reality and virtual reality where indoor scene segmentation is paramount – another task which relies on a highly diverse input distribution.    

Under the umbrella of bias removal, the taxonomy forms three natural groupings:

\begin{itemize}

\item Those seeking to increase generalisation and thus reduce the effects of bias via image augmentation. 

\item Those using the network architecture to attempt to remove or mitigate a known bias.

\item Those attempting to learn the bias within a given dataset and mitigate it accordingly.

\end{itemize}

Image augmentation increases variability in the training data by adding variation to images. This synthetically increases the training set without affecting the information contained.  Common perturbations include crops, sheers, flips, and colour jitter. This technique has been well researched in computer vision tasks, specifically image classification \cite{hendrycks2020, zhang2021a}.

Augmentation techniques have been adapted specifically to semantic segmentation where sheering and flipping images may not be the most appropriate approach. Kamann et al. \cite{kamann2020} propose the use of a colour mask gained from alpha-blending the ground truth segmentation map with the input data during training, which they coin \enquote{Painting-by-Numbers}. Building on the evidence of Geirhos et al. \cite{geirhos2019} who showed that CNNs are biased towards texture, Painting-by-numbers improves the robustness of semantic segmentation models to common image corruptions by making the texture of image classes less reliable and pushes the model to use geometry in the image to perform effective segmentation.  This technique does not require more training data and is thus efficient during training. Also motivated by CNN textural reliance, Jackson et al. \cite{jackson} explore style randomisation via altering colour and texture of the input image using style transfer whilst preserving semantic content, again showing an increase in accuracy.   

Multi-head models have been explored \cite{kim2019, alvi2018, wang2019} posing the ability of networks to \emph{unlearn} a known bias. In fact, it has been shown in \cite{wang2019} that some models even exacerbate the biases that are known in the datasets after training is complete, thus, models are using the bias itself as a cue to make a certain categorisation. Kim et al. \cite{kim2019} demonstrate a proof-of-concept approach showing that after planting a synthetic colour bias in the MNIST dataset, the model uses the obvious colour cue for categorisation rather than geometry of the numbers. A second model head employs an iterative algorithm using reverse gradients to successfully \enquote{unlearn} the \emph{known} colour bias, pushing the model to use the shape of the numbers as the cue for correct classification. To the best of our knowledge, such a technique has not been applied to highly variable domains such as semantic segmentation. As such, in this paper, we attempt to evaluate its success with semantic segmentation.       

The contribution of this paper is to increase the robustness of CNNs via the mitigation of algorithmic bias - specifically colour bias found in highly variable urban road scenes. Building on the foundations of Kim et al. \cite{kim2019}, we aim to implement an \enquote{unlearning} procedure within the network architecture itself rather than increase generalisability via augmentation of the input data. The unlearning procedure employs a multi-headed network to adversarially remove a target bias using reverse gradient loss. Semantic segmentation is used as a vehicle to assess the effectiveness of such a system, albeit the core principle should, in theory, be applicable to any deep learning architecture.   

\section{Removing a Known Bias}
\label{section:removing-bias}

This work takes advantage of the work in \cite{kim2019} entitled \emph{Learning Not to Learn} referred to, hereafter, as \emph{LNTL}. This proof-of-concept paper synthesises colour bias into the MNIST \cite{lecun1998} dataset and successfully uses a gradient reversal strategy to remove the colour information from the training data. In real data, we are not afforded the luxury of knowing exactly what the bias is and where it manifests -- although it has implicitly been shown that CNNs can pick up on the wrong cues \cite{geirhos2019}. Thus, a comparison between standard training data, a greyscale baseline and an implementation of bias \emph{unlearning} is tested. This strategy is a self-supervised task as we can extract the true colour labels from pixel values of the training data, which we already have. 

\vspace*{1mm}

\begin{figure*}[t!]
	\centering
	\includegraphics[width=0.95\linewidth]{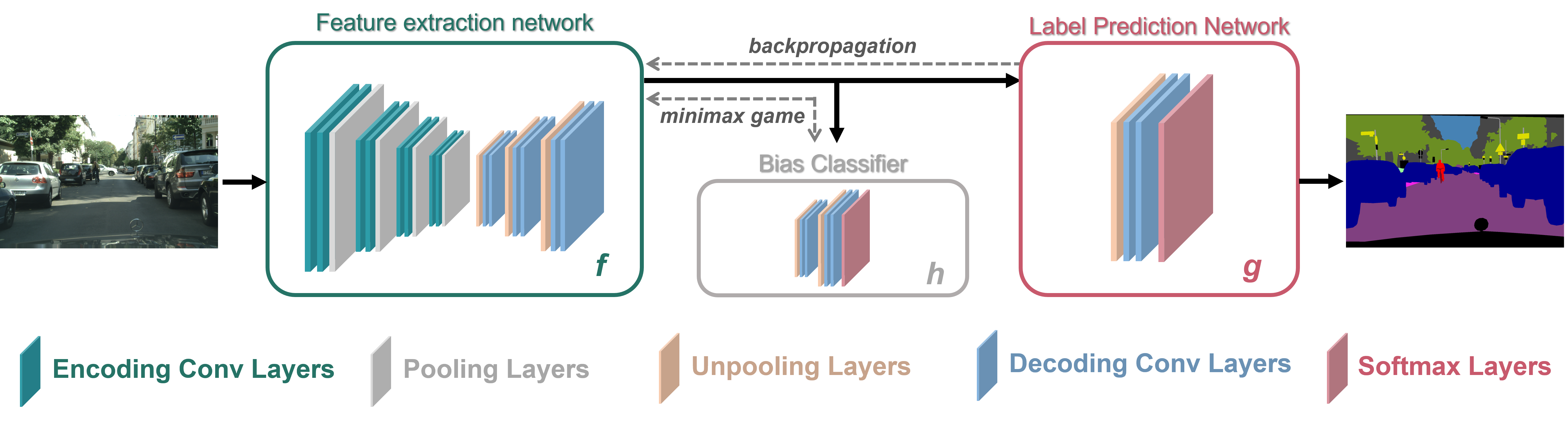}
	\captionsetup[figure]{skip=7pt}
	\captionof{figure}{Network architecture showing separate networks \emph{f}, \emph{g} and \emph{h} and their roles in the system.}
	\label{fig:network}\vspace{-0.25cm}
\end{figure*}

\subsection{Problem Statement}
\label{section:problem-statement}
\vspace*{-0.5mm}

In theory, the set of all images that self-driving cars encounter at test time is drawn from one set. This set contains all possible situations that the car could ever encounter, whether it be across a desert track in midday sun or a snow storm in a mountain pass. Indeed, many landscapes could provide this challenge in a single journey - confirming that this rich theoretical variation is not hyperbole. Practicality dictates that the training dataset is a restricted subset, and due to this, it is not a true representation of situations we \emph{could} encounter in the real world. In this sense, we assume the test set to be unbiased. We aim to train a network on biased data, attempt to systematically \emph{unlearn} that bias during the training phase and then deploy the model to perform on unseen and unbiased test data. 

Training and test sets are normally split randomly, in an attempt to eliminate the domain gap between the two distributions $X_{Train}$ and $X_{Test}$. Despite best efforts, a gap can still remain, and an even larger gap is the domain shift between $X_{Test}$ and the actual distribution $X$. For this reason, the results of the bias mitigation strategy might appear slightly subdued. Analogously, results published for segmentation accuracy are inflated compared to performance in the real world. The Cityscapes dataset \cite{cordts2016} - which consists of dash-cam footage from European cities - is partitioned on city for train, validation, and test sets, which means that model may lean on the nuances apparent in the training cities, making it an ideal candidate to use for the purposes of this work. 

Our setup splits a standard semantic segmentation classification network (e.g. DeepLabV3), into a double-headed network -- one with a pixel-wise classification head for the task of semantic segmentation and one with an auxiliary bias classification head. The networks are modular constructs consisting of a feature extractor network, $f: \mathcal{X} \mapsto \mathbb{R^N}$, a pixel-wise classification network, $g: \mathbb{R^N} \mapsto \mathcal{Y}$, and the bias classification head, $h: \mathbb{R^N} \mapsto \mathcal{B}$, where $N$ is the amount of feature maps produced by the embedding network $f$. 

Figure \ref{fig:network} shows the implemented network architecture, with the sub-networks $f$, $g$ and $h$, with the fork depicted at the last convolutional layer prior to classification. The precise architecture is left intentionally vague because the theory should apply to any system. Specific networks are discussed in more detail in Section \ref{subsection:networks}. 

\subsection{A Caveat on Fork Placement}
\label{subsection:fork}
\vspace*{-0.5mm}

Interestingly, since $f$ feeds its output feature maps into network $g$, the bias propagates through the network, and so the location of the fork is an arbitrary choice. Furthermore, $f \circ g$ will be void of any bias so long as $h$ has done its job in correctly classifying the bias and the subsequent gradient reversal step successfully discourages $f$ from using such cues. Prior work \cite{simonyan2014, yosinski2015} has shown that feature maps extracted at the start of a CNN generally contain low-level information, often containing blocks of colour and edges, whereas the final layer feature maps contain higher-level features with tightly integrated colour information.  

Intuitively, we expect that the feature maps towards the end of the network would be the most suitable place to add the fork. The training regime requires a short burst of end-to-end training without including the bias classification network, $h$. This ensures that the network already has some classifying ability and weights are converging. If we do not allow this head start for the classifying network, when the auxiliary bias network is activated, a mode collapse situation could occur, where the network weights are unable to converge towards optimisation \cite{kim2019}.

Thus, a fork located towards the end of the main bulk of the segmentation network would allow the weights upstream to be amended whilst the classification layer would be largely unaffected. We hypothesise that a different fork location further upstream would achieve the same result eventually but would take longer to reach convergence. Furthermore, the fork has been added leaving one convolutional layer before the SoftMax layer. This ensures learnable parameters remain in the classification head and allows for any reactive adjustments in network $g$ from a change of its input, $f(x)$.
 

\section{Experiments}
\label{sec:experiments}

The following section details the experiments undertaken, and provides results and analysis. As such, in this section, the datasets are introduced (Section \ref{subsection:datasets}), an explanation of the specific networks used are provided (Section \ref{subsection:networks}), evaluation metrics are explained (Section \ref{subsection:metrics}) and all experiments and results are discussed. 

\subsection{Datasets}
\label{subsection:datasets}

\textbf{Cityscapes} \cite{cordts2016} $\>\>\>$ The Cityscapes dataset is a public and widely-used semantic segmentation benchmark. Cityscapes contains data from 50 cities and images are annotated with 30 semantic classes. The dataset contains over 20,000 coarsely annotated images and $\approx$ 5,000 finely annotated images providing pixel-level, instance-level and panoptic semantic ground truth labels. Raw images and segmentation masks are provided in portable network graphics format. Data contains both 16-bit High-Dynamic Range images and 8-bit Low-Dynamic Range format to use at an image size of 1024$\times$2048. We evaluate our semantic segmentation performance on the official 500 annotated validation set.  

\textbf{SYNTHIA} \cite{ros2016} $\>\>\>$ The SYNTHIA dataset is also publicly available and comprised of nearly-photo-realistic images from a synthetically-rendered virtual city. The dataset used in this paper is the \emph{synthia-rand-cityscapes} subset containing 9,400 images at a resolution of 1280 $\times$ 760, which contains labels compatible with Cityscapes, allowing for a fairer examination of the results. The dataset provides fine detail instance segmentation labels - thus data preprocessing is undertaken to create semantic labels from the data provided.

SYNTHIA images contain very realistic granular detail and complex scenes; some images contain very high numbers of pedestrians - all of which have pixel perfect annotation, leveraging the automatic ability to label images in a generated scene. Further, SYNTHIA has a rich variety of luminance, high scene diversity and vantage differences, making it a more variable sample space than Cityscapes. Input perturbations during rendering create similar images with slight nuances. When creating 70\% / 30\% training / validation sets, the data is split as though a temporal dependence exists within the data. This ensures that similar images do not occur in both the training and validation sets -- providing a more representative interpretation of a real setting, where a test set is a wholly unseen set of images. This also ensures that we do not get over-optimistic metrics upon evaluation.   

\subsection{Network Architecture}
\label{subsection:networks}

Since the network $f$ consists of a feature extraction sub-network, we have the flexibility to choose any architecture we wish. Indeed, we have positioned this paper for the task of mitigating colour bias in urban scenes, but we could equally apply the technique to other fields of computer vision; say, facial recognition de-biasing. 

We have chosen to test two seminal semantic segmentation architectures, namely, DeeplabV3 and SegNet as discussed in Section~\ref{section:related-work}. DeeplabV3 is trained with multiple ResNet backbones with ImageNet \cite{deng2009} pretrained weights. Adopting this pretraining procedure could add noise to the results as it adds uncertainty about the origins of the CNN bias. Nevertheless, it is an efficiency trade off, and as mentioned in Section~\ref{subsection:fork}, anything upstream from the fork can be unlearned. The fork is located directly after the concatenation of the ASPP module within DeepLabV3 and $f(X)$ outputs $1280$ feature maps to both the auxiliary bias head, $h$, and the primary semantic classifier, $g$.   

SegNet follows a simple symmetric encoder-decoder architecture. The encoder is topologically identical to the convolutional layers of the VGG16 \cite{simonyan2015} network, whilst the decoder upsamples hierarchically by using the indices of the corresponding max pooling operation from the encoding operation. Again, the fork for network $h$ is placed before the last convolutional layer and $f(X)$ outputs $64$ feature maps to both $h$ and $g$.

\subsection{Metrics}
\label{subsection:metrics}

\begin{figure}[t!]
	\centering
	\includegraphics[width=0.99\linewidth]{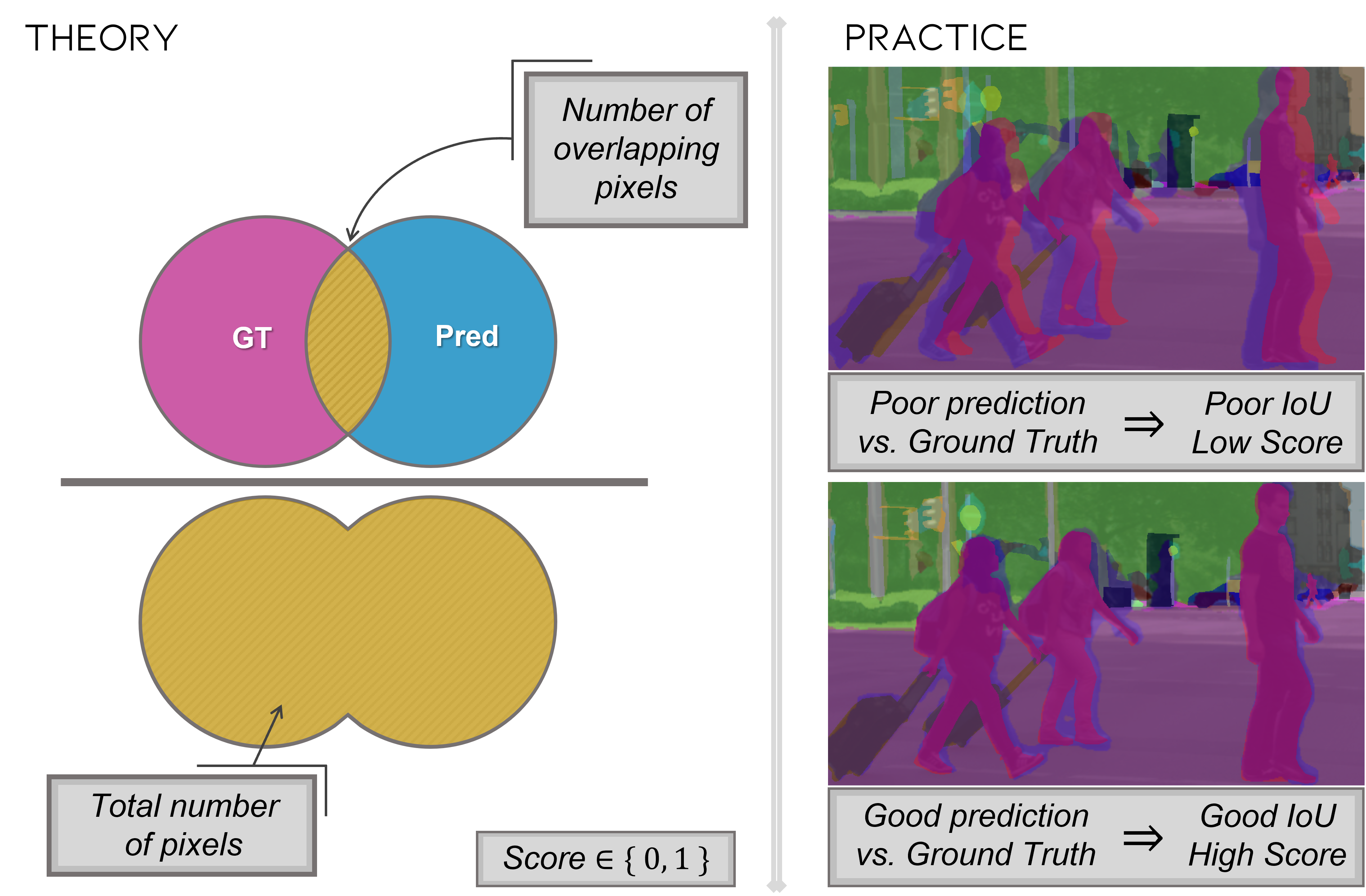}
	\captionsetup[figure]{skip=7pt}
	\captionof{figure}{Left pane: Visual demonstration of intersection over union calculation. Right Pane: Predicted segmentation masks from our trained DeepLabV3 model are overlaid on ground truth image masks to demonstrate realistic IOU scores.}
	\label{fig:metrics}\vspace{-1mm}
\end{figure}

In semantic segmentation dealing with multi-class problems, we often encounter the issue of class imbalance. This occurs when background parts of an image, say buildings or sky, dominate the total pixel count compared to say that of a pedestrian or a red traffic light. Due to this, accuracy becomes an ambiguous metric; inaccuracy of minority classes gets overshadowed by the accuracy of majority classes. Furthermore, in the context of driving, the more infrequent classes are often the most important for human safety.  As a result, it is common to use the intersection over union metric (IoU). IoU measures the ratio between the amount of overlap between the predicted and ground truth pixels and the total number of pixels taken up by the prediction and the ground truth. See Figure \ref{fig:metrics} for a visual interpretation.  

It is customary to use the mean intersection over union (mIoU) which is quoted in this paper. However, during evaluation, we also compute and inspect the individual IoU per category to give a more granular understanding of the model performance. Scores theoretically fall between 0 and 1, although percentages are often quoted. 

\subsection{Comparing the Baseline and LNTL Schemes}
\label{subsection:baseline-vs-lntl}

All models are trained for 100 epochs until convergence. A learning rate of $0.001$ is used with the Adam optimiser \cite{kingma2014adam}. Learning rate decay is enforced with the scheduler reducing the learning rate by a factor of $0.1$ every 40 epochs. Class weights are computed for all the training data so the cross-entropy loss function could allow for class imbalance, a common occurrence in urban driving scenarios. 

Baseline models are trained for both SegNet and DeepLab, with inputs of colour training images and greyscale training images. Since the LNTL scheme penalises the classifier by using the gradient reversal module to adjust the weights, it was expected that the accuracy may suffer somewhat. We did, however, expect that the LNTL scheme would perform better than the networks making predictions using limited colour information, as is supplied in the greyscale training set.

\begin{figure}[h!]
\centering
\begin{minipage}{0.99\linewidth}
  \centering
\includegraphics[width=0.98\linewidth]{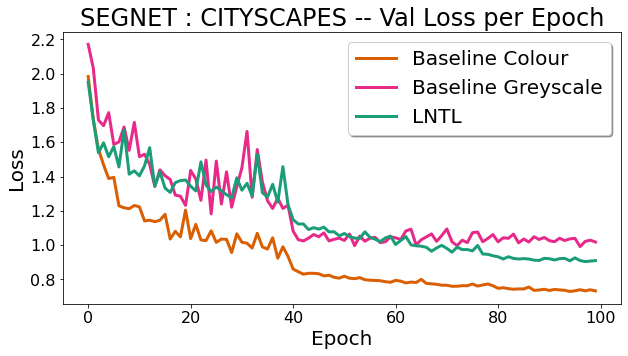}
\end{minipage}
\vspace*{0.8mm}
\begin{minipage}{0.99\linewidth}
  \centering
\includegraphics[width=0.98\linewidth]{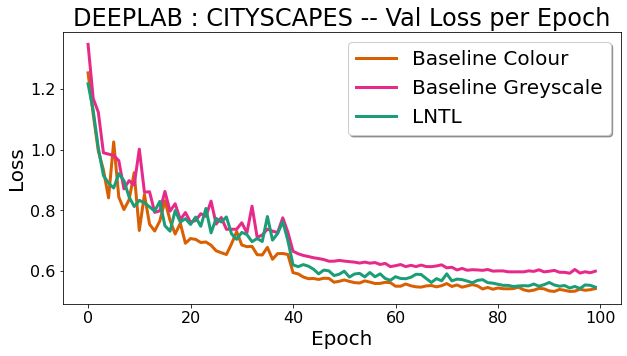}
\end{minipage}%

\caption{Learning Not To Learn scheme vs. colour and greyscale baseline for SegNet and DeepLab segmentation architectures.}
\label{fig:first_losses}
\end{figure}

This intuition is confirmed given the loss curves displayed in Figure \ref{fig:first_losses}; the LNTL scheme has penalised the loss compared to the converged colour input data for both SegNet and DeepLab architectures. Note that the SegNet minimum loss is $0.730$, which DeepLab achieves after just 20 epochs of training. Of course, further image augmentation, extra training data and hyperparameter tuning could be used to drive down the loss even further and improve accuracy; however, to satisfy the project hypothesis, we only need a comparable canvas to test the concept of applying colour unlearning within the domain of urban scenes. In subsequent experiments, the Deeplab network is favoured for its slightly more stable learning, and higher accuracy.

\subsection{Synthesising a Covariate Shift in Validation Data}

In order to properly test our hypothesis that the LNTL scheme can unlearn colour information in highly variable scenes such as urban images, it is necessary to test both the baseline DeepLab architecture and the LNTL scheme on unseen data which contains a covariate shift from images it was trained on. To this end, three synthesised validation sets are created:

\vspace{1mm}

\begin{itemize}

\item converting the validation set to greyscale,

\item applying random colour jitter to validation images,

\item applying a colour invert transform to validation images.

\end{itemize}

\vspace{2mm}

\begin{figure}[h!]
	\centering
	\includegraphics[width=0.98\linewidth]{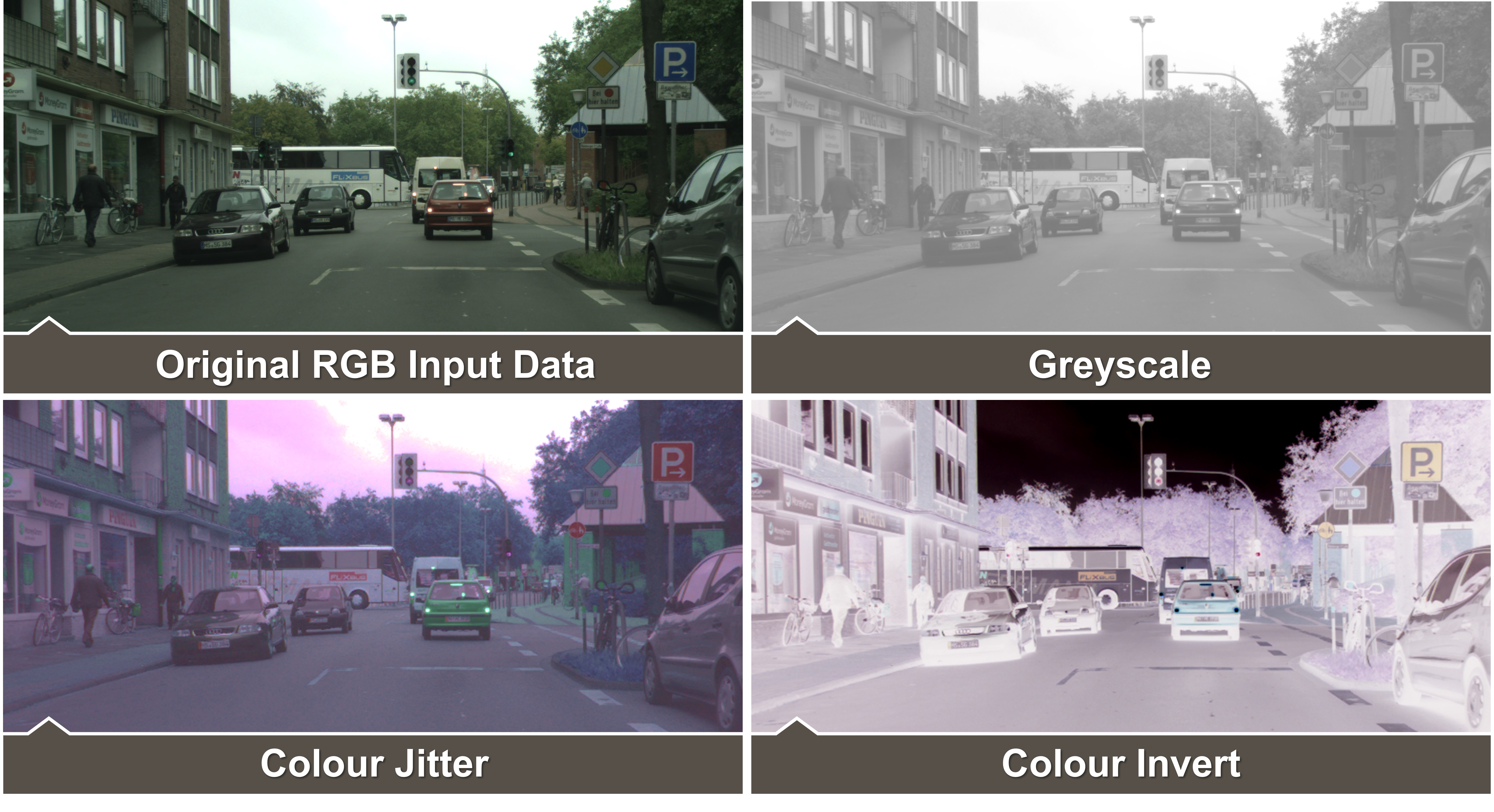}
	\captionsetup[figure]{skip=7pt}
	\captionof{figure}{Input image manipulations.}
	\label{fig:manipulations}\vspace{-3mm}
\end{figure}

Visual examples of these transformations are shown in Figure \ref{fig:manipulations}. Models are re-trained on the normal training set, both on SYNTHIA and Cityscapes datasets and only validated on the transformed images. All other parameters remain the same to allow for a fairer comparison. This enables us to monitor the network loss over the training cycle to see if the LNTL scheme slowly improves in its validation convergence. Monitoring loss in this way allows us to assess overall model health, since the network is technically focussed on \emph{minimising} loss not \emph{maximising} accuracy. Furthermore, during training, we can monitor the loss in the bias head to check for signs of divergence. This behaviour is welcomed and can be an indicator that colour information is being extracted out of the feature maps in network $f$. This makes categorisation in the bias head more difficult thus manifesting as a diverging loss in network $h$.

Intuitively, it is worth noting that although greyscale images remove a lot of the colour information, distances between pixel values can still be leveraged, synonymously for colour jitter, despite the stochasticity. Colour invert, however, maximises this difference and corrupts this relationship the most, disrupting spatial inter-correlations between different pixel values. 

\begin{figure*}[h!]
\centering
\begin{minipage}{0.98\textwidth}
  \centering
\includegraphics[width=0.9\textwidth]{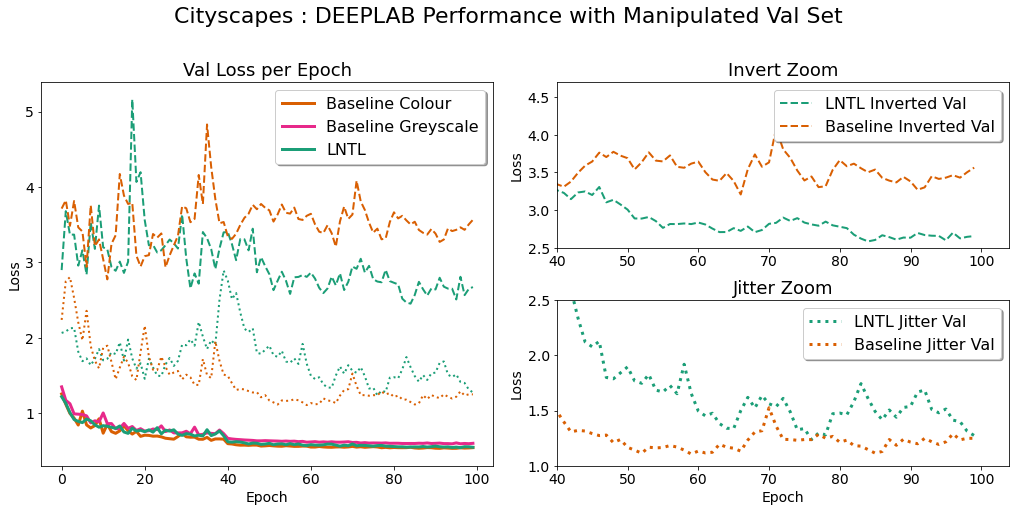}
\end{minipage}%

\begin{minipage}{0.98\textwidth}
  \centering
\includegraphics[width=0.9\textwidth]{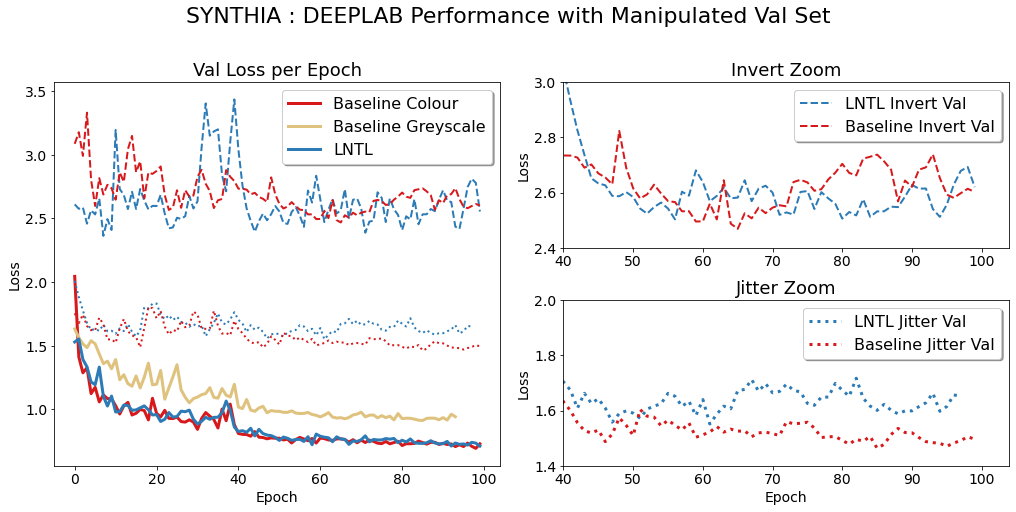}
\end{minipage}%

\captionsetup[figure]{skip=7pt}
\captionof{figure}{Top Pane: Cityscapes dataset. Bottom Pane: SYNTHIA dataset. Large graph shows the validation results on the normal validation set and the results for manipulated validation sets with colour invert (dashed) and colour jitter (dotted). Adjacent are magnified and truncated plots to the last 40 epochs for the manipulations.}
\label{fig:losses_city_synth}
\end{figure*}

Figure \ref{fig:losses_city_synth} shows the disparity in loss between uncorrupted validation images and those with perturbed colour. Reference lines on the bottom of the graph highlight the extent to which the networks overfit to the colour in the training data. In the Cityscapes dataset, the LNTL scheme shows a clear improvement over the baseline method when colour invert is applied to the validation images. This suggests that the LNTL scheme has increased model robustness and has diminished its dependence on colour for categorisation. Although random colour jitter in Cityscapes validation has an increased loss, it appears to be converging on a downward trajectory.  

Results are not as desirable for the SYNTHIA dataset. Firstly, we notice a significant reduction in disparity between uncorrupted and corrupted validation images. This may be due to the rich variety of luminance in SYNTHIA, as mentioned in Section~\ref{subsection:datasets}. Colour invert seems to affect both schemes equally, whilst colour jitter is marginally favourable to the baseline scheme. As we may expect, jitter creates less of an impact on the loss than the invert in all cases.

\subsection{Using Synthesised Weather Corruptions as a Proxy for Different Driving Conditions} 

From the Cityscapes data repository, researchers \cite{hu2019, sakaridis2018} have created imitations of rain and fog over the normal Cityscapes training data. Ground truth labels are exactly the same as for the standard training data, so we can leverage this dataset to more closely resemble test images an autonomous vehicle may encounter in the wild. Different severities of image manipulation are provided. We randomly select one level of severity for each image, yielding a 295-image validation set for rain and 550-image validation set for fog.  

\begin{figure}[H]
	\centering
	\includegraphics[width=0.98\linewidth]{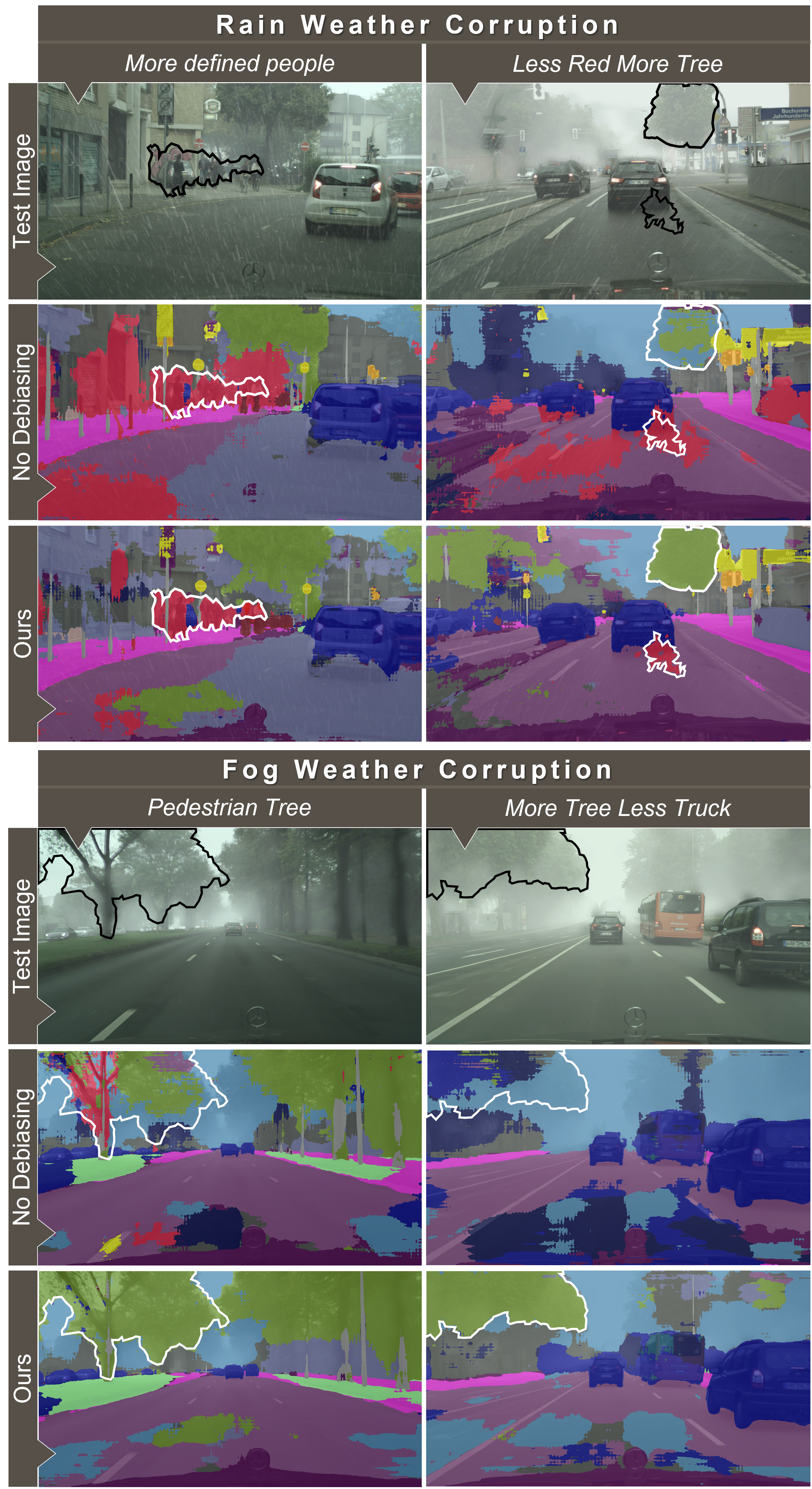}
	\caption{Qualitative results from running the best models displayed in Figure \ref{fig:first_losses} for the baseline and the LNTL models. Regions of interest are shown in boundaries.} 
	\label{fig:rain_fog}\vspace{-0.01cm}
\end{figure}

Figure \ref{fig:rain_fog} uses the synthesised rain and fog validation images fed into the best trained model for each of the baseline and LNTL schemes. All prediction images contain more noise and demonstrate much poorer performance, as expected from the loss curves in Figure \ref{fig:losses_city_synth}. Bounding boxes, denoting regions of interest, show that the LNTL scheme has managed to better segment the pedestrians in the first quadrant scene. The baseline model has produced nonsensical predictions, hallucinating pedestrians on the building in the scene. In the second quadrant \enquote{\emph{Less Red More Tree}} we observe similar erroneous red patches - our debiasing method mitigates much of this and correctly identifies more tree through the synthesised fog. Similar pedestrian hallucinations can be seen in the tree in the fog in the third quadrant, and again the de-biasing scheme is more accurately able to segment the tree through the fog. Once more the fourth quadrant demonstrates the ability of the bias unlearning network to correctly classify a tree given a corrupted test image, where comparatively the baseline struggles. The fog image corruptions perform better than the baseline on average (\emph{see Table \ref{table:av_miou}}). Although both predictions contain noise and artefacts, results show a robustness to covariate shift at test time for the bias unlearning scheme. Table \ref{table:categories} shows congruence with qualitative observations with a clear improvement in the \emph{Nature}, \emph{Human} and \emph{Vehicle} categories.

\subsection{Quantitative Results}

A distinction between classes and categories is to be made; classes comprise the objects the model is seeking to predict used in the Softmax function of the network. In this case, we adopt the 20 Cityscape classes consisting of \emph{Road}, \emph{Sky}, \emph{Car}, \emph{Truck}, \emph{Person}, \emph{Rider}, etc. On the other hand, categories consist of groupings of classes -- e.g. \emph{Human}, \emph{Vehicle}, \emph{Nature} and so on. Table \ref{table:av_miou} provides average class-wise mIoU scores for both the DeepLabV3 baseline model and our bias unlearning model. Bold values highlight the best performers for each image transformation. The LNTL scheme not only performs marginally better on the original image, but it performs consistently better when validated on an out-of-distribution test image - only failing to beat the baseline in the \emph{rain} validation set. This could be due to the relatively small size of the rain validation set compared to others tested, hindering model convergence.

The disparity between the manipulated validation sets and the RGB images empirically shows just how severe the lack of generalisability is - these results come from the same images only with colour corruptions added. This  demonstrates that the networks \emph{do} overfit to colour information available, highlighting the Occam's Razor nature of CNN learning, that is if the colour information is sufficient to drive down the loss, then why use another cue? Indeed, the results allow to quantify this degree of overfitting with an average reduction of 41.80\% in mIoU score over all validation sets reported. The worst observed covariate shift is from RGB $\to$ Invert giving a mIoU reduction of 85.50\%.

{\renewcommand{\arraystretch}{1.2}
\begin{table}
\centering
\caption{Class-wise averages of minimum loss and mIoU validation results for a DeepLabV3 model trained on Cityscapes training set. Bold typeface indicates optimal results.}
\label{table:av_miou}
\begin{tabu}{@{\extracolsep{2.5pt}}c | l | c c c c@{}} 
\cline{3-6} \cline{3-6}
\multicolumn{1}{c}{} & \multicolumn{1}{c}{} & \multicolumn{2}{c}{\textbf{mIoU (\%)}} & \multicolumn{2}{c}{\textbf{loss}}  \\
\cline{3-4} \cline{5-6}
\multicolumn{1}{c}{} & \multicolumn{1}{c}{}  & Baseline & \emph{Ours} & Baseline & \emph{Ours} \\
\hline\hline
Original  & RGB & 58.50  & \textbf{58.80} & \textbf{0.542} & 0.546 \\
\hline
\multirow{3}{1.8cm}{\centering Image Manipulation} & Greyscale & 36.20 & \textbf{37.50} & 1.173 & \textbf{1.156} \\
 & Invert & 8.50 & \textbf{13.70} & 3.112 & \textbf{2.423} \\
 & Jitter & 33.30 & \textbf{34.10} & 1.259 & \textbf{1.180} \\
\hline
\multirow{2}{1.8cm}{\centering Weather Corruption} & Rain & \textbf{39.40} & 38.90 & \textbf{1.190} & 1.249 \\
 & Fog & 52.80 & \textbf{54.00} & 0.873 & \textbf{0.788} \\
\hline \hline

\end{tabu} 
\end{table}


{\renewcommand{\arraystretch}{1.2}
\begin{table*}
\centering
\caption{Category-wise validation mIoU results for a DeepLabV3 model trained on Cityscapes training data. Observations of interest are highlighted in bold and discussed.}
\vspace*{2mm}
\label{table:categories}
\begin{tabu}{l | @{\extracolsep{1.5mm}} c c c c c c c c c c c c @{}} 
\hline\hline
\multirow{2}{*}{Categories} & \multicolumn{2}{c}{Normal} & \multicolumn{2}{c}{Invert} & \multicolumn{2}{c}{Jitter} & \multicolumn{2}{c}{Greyscale} & \multicolumn{2}{c}{Rain} & \multicolumn{2}{c}{Fog} \\ 
\cline{2-3} \cline{4-5} \cline{6-7} \cline{8-9} \cline{10-11}  \cline{12-13}
      & \emph{Baseline}	& \emph{Ours} & \emph{Baseline}	& \emph{Ours} & \emph{Baseline}	& \emph{Ours} & \emph{Baseline}	& \emph{Ours} & \emph{Baseline}	& \emph{Ours} &\emph{Baseline}	& \emph{Ours} \\ 
\hline\hline
Flat      		& 93.6 & 93.1 & 32.8 & 55.8 & 80.3 & 74.6 & 88.7 & 85.7 & 71.9 & 55.1 & 94.3 & 92.3  \\
Construction    & 85.0 & 85.4 & 20.5 & 40.5 & 56.6 & 64.1 & 61.9 & 64.4 & 55.6 & 47.5 & 72.6 & 73.7  \\
Object      	& 47.7 & 49.0 & 9.8  & 14.7 & 27.3 & 28.6 & 31.1 & 30.8 & 34.7 & 38.1 & 41.5 & 43.4  \\
Nature      	& 87.6 & 87.5 & 3.6  & \textbf{34.6} & 68.7 & \textbf{72.0} & 54.8 & \textbf{56.2} & 49.0 & \textbf{58.5} & 52.8 & \textbf{58.1}  \\
Sky      		& 86.8 & 86.0 & 0.3  & 2.6  & 65.1 & 62.3 & 78.7 & 71.9 & 63.4 & 69.0 & 55.1 & 56.4  \\
Human      		& 66.5 & \textbf{67.8} & 13.9 & 10.1 & 19.3 & \textbf{31.8} & 29.8 & \textbf{39.4} & 48.5 & \textbf{55.6} & 63.3 & \textbf{64.6}  \\
Vehicle      	& 86.3 & \textbf{86.8} & 17.0 & \textbf{26.9} & 38.6 & \textbf{49.8} & 49.5 & \textbf{64.5} & 46.9 & \textbf{47.5} & 80.8 & \textbf{81.5}  \\
\hline\hline
Average           & 79.1 & 79.4 & 14.0  & 26.5 & 50.8 & 54.8 & 56.4 & 59.0 & 53.2 & 53.0 & 65.8 & 67.1  \\
\hline 
\emph{Ours} $\pm$\%     & 	  & +0.3 &      & +89.3 &     & +7.9 & 	   & +4.6 &      & -0.3 &      & +2.0  \\
\hline\hline
\end{tabu}
\end{table*}

Table \ref{table:categories} shows the category-wise intersection over union scores providing a more granular understanding of the performance of the networks. The penultimate two rows display the average class scores and the performance increase, or decrease observed by the LNTL from the baseline. These values, and all others quoted in this paper, are displayed as a \textit{percentage increase} from one percentage to the other, not simply a difference in percentages. Consistency is maintained in reporting value differences throughout our work.     

Bold values in Table \ref{table:categories} highlight some interesting observations from these experiments. Firstly, the LNTL scheme performs consistently better in the \enquote{human}, \enquote{nature} and \enquote{vehicle} classes than the baseline model. One subjective interpretation is that these categories in particular have a more unique, and largely unchanged geometry from scene to scene, i.e. the human form is distinctive and largely unchanged from individual to individual. Furthermore, the ASPP module of DeepLab handles objects at different scales. In contrast, the category-wise results show the model performing consistently worse at predicting the \enquote{flat} class - perhaps, in the same vein, from buildings having no fixed geometry from scene to scene. This could mean that we have coerced the network to use more geometric-based cues to perform classification rather than colour. Quantitative results agree with qualitative observations. Furthermore, the debiasing scheme performs unanimously better in the \enquote{human} category than the baseline architecture. This is a promising result, given the safety-critical nature of the applications of segmentation technology.

\section{Limitations and Future Work}
\label{section:future-work}

Further work is needed to reinforce the reported findings. A deeper analysis of class-wise mIoU scores would provide more insight into precisely where the bias manifests within images. In particular, more granular understanding of the class-wise false-positives and false-negatives of predictions would be useful since in safety-critical applications such as autonomous vehicles, this is a vital requisite. This means that failing to correctly identify a pedestrian has attached with it more gravity than incorrectly classifying a pole, moreover - classifying a pedestrian as \enquote{road} has more consequence than classifying a pedestrian as a \enquote{rider}. 

Another consideration is analysing the extent to which augmentation techniques, e.g. \cite{kamann2020}, interact with the proposed bias unlearning scheme. Does the robustness offered by adequate augmentation reduce the performance observed in this work - or does it compliment it? In addition, urban scene data has high temporal dependence in the wild - a domain of active research and one which would be interesting to incorporate within our scheme.

Additionally, this project only focused on the mitigation of a known bias - colour. In the burgeoning world of big data, it is often unsurmountable to assess data for such bias. Furthermore, we are also at the mercy of our own biased representations. Amini et al. \cite{amini2019} tackle this issue with the use of variational autoencoders (VAEs). The proposed model actually learns, in an unsupervised manner, the latent structure of the input data and adaptively uses this learned latent distribution to selectively upsample underrepresented data points. This allows the model itself to determine the bias inherent within the data during training. Although Amini et al. demonstrate this technique through racial and gender bias in facial recognition systems, the idea itself is generalisable to multiple domains. We have shown that while colour bias does exist, the inter-correlation between variables in the input distribution may be more sophisticated than simply penalising a colour value.

\section{Conclusion}
\label{section:conclusion}

In this paper, we applied a colour bias unlearning scheme to highly variable images of urban road scenes as an iterative learning process during training. Our contribution empirically shows that semantic segmentation architectures do overfit to the colour within training data, and they struggle to generalise to unseen test data -- even from a very similar input distribution, as seen in raw $\to$ weather manipulation experiments. In the worst case, when validating on a set with a colour invert transformation, reductions of 85.50\% were observed.  We demonstrate that the \emph{unlearning} technique itself is viable, showing a qualitative improvement to both \emph{stuff} and \emph{things} classes in pixel-wise semantic segmentation, from a benchmark seminal architecture - mIoU metrics confirm this improvement. We observed a 62\% increase in mIoU score for colour invert; when neglecting the result for colour invert, we still observe an average increase of 1.5\% over all validation set manipulations tested. Furthermore, an average increase of 14.5\% is observed for the \enquote{human} class, enhancing pragmatic performance in a safety-critical application such as autonomous driving. We position this paper to push towards robust, trustworthy technology - aiming for a transparent and explainable future in artificial intelligence, alleviating algorithmic bias.


\bibliographystyle{IEEEtran}
\bibliography{custom}

\begin{thebibliography}{10}
\providecommand{\url}[1]{#1}
\csname url@samestyle\endcsname
\providecommand{\newblock}{\relax}
\providecommand{\bibinfo}[2]{#2}
\providecommand{\BIBentrySTDinterwordspacing}{\spaceskip=0pt\relax}
\providecommand{\BIBentryALTinterwordstretchfactor}{4}
\providecommand{\BIBentryALTinterwordspacing}{\spaceskip=\fontdimen2\font plus
\BIBentryALTinterwordstretchfactor\fontdimen3\font minus
  \fontdimen4\font\relax}
\providecommand{\BIBforeignlanguage}[2]{{%
\expandafter\ifx\csname l@#1\endcsname\relax
\typeout{** WARNING: IEEEtran.bst: No hyphenation pattern has been}%
\typeout{** loaded for the language `#1'. Using the pattern for}%
\typeout{** the default language instead.}%
\else
\language=\csname l@#1\endcsname
\fi
#2}}
\providecommand{\BIBdecl}{\relax}
\BIBdecl

\bibitem{girshick2013}
R.~Girshick, J.~Donahue, T.~Darrell, and J.~Malik, ``{Rich Feature Hierarchies
  for Accurate Object Detection and Semantic Segmentation},'' in \emph{IEEE
  Computer Society Conf. Computer Vision and Pattern Recognition}, 2013.

\bibitem{atapour2018real}
A.~Atapour-Abarghouei and T.~P. Breckon, ``Real-{{Time Monocular Depth
  Estimation Using Synthetic Data}} with {{Domain Adaptation}} via {{Image
  Style Transfer}},'' in \emph{{IEEE/CVF Conf.} Computer Vision and Pattern
  Recognition}, 2018, pp. 2800--2810.

\bibitem{devlin2018bert}
J.~Devlin, M.-W. Chang, K.~Lee, and K.~Toutanova, ``{Bert: Pre-training of Deep
  Bidirectional Transformers for Language Understanding},'' \emph{arXiv
  preprint arXiv:1810.04805}, 2018.

\bibitem{atapour2019king}
A.~Atapour-Abarghouei, S.~Bonner, and A.~S. McGough, ``{A King’s Ransom for
  Encryption}: {Ransomware} {Classification using Augmented One-Shot Learning}
  and {Bayesian} approximation,'' in \emph{IEEE Int. Conf. Big Data}.\hskip 1em
  plus 0.5em minus 0.4em\relax IEEE, 2019, pp. 1601--1606.

\bibitem{badrinarayanan2017}
V.~Badrinarayanan, A.~Kendall, and R.~Cipolla, ``{{SegNet}}: {{A Deep
  Convolutional Encoder}}-{{Decoder Architecture}} for {{Image
  Segmentation}},'' in \emph{IEEE Trans. Pattern Analysis and Machine
  Intelligence}, vol.~39, no.~12, 2017, pp. 2481--2495.

\bibitem{atapour2020rank}
A.~Atapour-Abarghouei, S.~Bonner, and A.~S. McGough, ``{Rank over Class}: {The
  Untapped Potential of Ranking in Natural Language Processing},'' \emph{arXiv
  preprint arXiv:2009.05160}, 2020.

\bibitem{chen2018}
L.-C. Chen, Y.~Zhu, G.~Papandreou, F.~Schroff, and H.~Adam,
  ``Encoder-{{Decoder}} with {{Atrous Separable Convolution}} for {{Semantic
  Image Segmentation}},'' in \emph{Euro. Conf. Computer Vision}, 2018, pp.
  833--851.

\bibitem{tan2020}
M.~Tan and Q.~V. Le, ``{{EfficientNet}}: {{Rethinking Model Scaling}} for
  {{Convolutional Neural Networks}},'' in \emph{Int. Conf. Machine Learning},
  2020, pp. 6105--6114.

\bibitem{kumar2020}
R.~P. C.~Kumar~B. and Mohana, ``{YOLOv3 and YOLOv4: Multiple Object Detection
  for Surveillance Applications},'' in \emph{Int. Conf. Smart Systems and
  Inventive Technology}, 2020, pp. 1316--1321.

\bibitem{lundberg2017}
S.~M. Lundberg and S.-I. Lee, ``{A Unified Approach to Interpreting Model
  Predictions},'' in \emph{Advances in Neural Information Processing Systems},
  2017.

\bibitem{fong2017}
R.~C. Fong and A.~Vedaldi, ``{Interpretable Explanations of Black Boxes by
  Meaningful Perturbation},'' in \emph{IEEE Int. Conf. Computer Vision}, 2017,
  pp. 3449--3457.

\bibitem{kirillov2019}
A.~Kirillov, K.~He, R.~Girshick, C.~Rother, and P.~Dollar, ``Panoptic
  {{Segmentation}},'' in \emph{{{IEEE}}/{{CVF Conf.}} {{Computer Vision}} and
  {{Pattern Recognition}}}, 2019, pp. 9404--9413.

\bibitem{hawke2021}
J.~Hawke, V.~Badrinarayanan, A.~Kendall \emph{et~al.}, ``{Reimagining an
  Autonomous Vehicle},'' \emph{arXiv preprint arXiv:2108.05805}, 2021.

\bibitem{chen2017b}
L.-C. Chen, G.~Papandreou, F.~Schroff, and H.~Adam, ``Rethinking {{Atrous
  Convolution}} for {{Semantic Image Segmentation}},'' in \emph{{{IEEE Conf.}}
  {{Computer Vision}} and {{Pattern Recognition}}}, 2017.

\bibitem{he2017}
K.~He, G.~Gkioxari, P.~Dollár, and R.~Girshick, ``{Mask R-CNN},'' in
  \emph{IEEE Int. Conf. Computer Vision}, 2017, pp. 2980--2988.

\bibitem{cheng2020}
B.~Cheng, M.~D. Collins, Y.~Zhu, T.~Liu, T.~S. Huang, H.~Adam, and L.-C. Chen,
  ``Panoptic-{{DeepLab}}: {{A Simple}}, {{Strong}}, and {{Fast Baseline}} for
  {{Bottom}}-{{Up Panoptic Segmentation}},'' in \emph{{{IEEE}}/{{CVF Conf.}}
  {{Computer Vision}} and {{Pattern Recognition}}}, 2020, pp. 12\,472--12\,482.

\bibitem{atapour2018extended}
A.~Atapour-Abarghouei and T.~P. Breckon, ``{Extended Patch Prioritization for
  Depth Filling within Constrained Exemplar-based RGB-D Image Completion},'' in
  \emph{Int. Conf. Image Analysis and Recognition}.\hskip 1em plus 0.5em minus
  0.4em\relax Springer, 2018, pp. 306--314.

\bibitem{papageorgiou1998general}
C.~P. Papageorgiou, M.~Oren, and T.~Poggio, ``{A General Framework for Object
  Detection},'' in \emph{Int. Conf. Computer Vision}.\hskip 1em plus 0.5em
  minus 0.4em\relax IEEE, 1998, pp. 555--562.

\bibitem{torralba2002depth}
A.~Torralba and A.~Oliva, ``{Depth Estimation from Image Structure},''
  \emph{IEEE Trans. Pattern Analysis and Machine Intelligence}, vol.~24, no.~9,
  pp. 1226--1238, 2002.

\bibitem{atapour2018comparative}
A.~Atapour-Abarghouei and T.~P. Breckon, ``{A Comparative Review of Plausible
  Hole Filling Strategies in the Context of Scene Depth Image Completion},''
  \emph{Computers \& Graphics}, vol.~72, pp. 39--58, 2018.

\bibitem{long}
J.~Long, E.~Shelhamer, and T.~Darrell, ``Fully {{Convolutional Networks}} for
  {{Semantic Segmentation}},'' in \emph{{{IEEE}}/{{CVF Conf.}} {{Computer
  Vision}} and {{Pattern Recognition}}}, 2015, p.~10.

\bibitem{ronneberger2015}
O.~Ronneberger, P.~Fischer, and T.~Brox, ``{U-Net}: {Convolutional Networks for
  Biomedical Image Segmentation},'' in \emph{Int. Conf. Medical Image Computing
  and Computer-Assisted Intervention}.\hskip 1em plus 0.5em minus 0.4em\relax
  Springer, 2015, pp. 234--241.

\bibitem{hyeonwoo2015}
H.~Noh, S.~Hong, and B.~Han, ``{Learning Deconvolution Network for Semantic
  Segmentation},'' in \emph{IEEE Int. Conf. Computer Vision}, 2015.

\bibitem{chen2016}
L.-C. Chen, G.~Papandreou, I.~Kokkinos, K.~Murphy, and A.~L. Yuille, ``Semantic
  {{Image Segmentation}} with {{Deep Convolutional Nets}} and {{Fully Connected
  CRFs}},'' in \emph{IEEE Trans. Pattern Analysis and Machine Intelligence},
  2016.

\bibitem{chen2017a}
L.-C. Chen, G.~Papandreou, I.~Kokkinos, K.~Murphy, and A.~Yuille,
  ``{{DeepLab}}: {{Semantic Image Segmentation}} with {{Deep Convolutional
  Nets}}, {{Atrous Convolution}}, and {{Fully Connected CRFs}},'' in
  \emph{{IEEE Trans. Pattern Analysis and Machine Intelligence}}, 2017, pp.
  834--848.

\bibitem{tao2020}
A.~Tao, K.~Sapra, and B.~Catanzaro, ``Hierarchical {{Multi}}-{{Scale
  Attention}} for {{Semantic Segmentation}},'' \emph{arXiv preprint
  arXiv:2005.10821}, 2020.

\bibitem{zhao2017}
H.~Zhao, J.~Shi, X.~Qi, X.~Wang, and J.~Jia, ``Pyramid {{Scene Parsing
  Network}},'' in \emph{{{IEEE}}/{{CVF Conf.}} {{Computer Vision}} and
  {{Pattern Recognition}}}, 2017.

\bibitem{cordts2016}
M.~Cordts, M.~Omran, S.~Ramos, T.~Rehfeld, M.~Enzweiler, R.~Benenson,
  U.~Franke, S.~Roth, and B.~Schiele, ``The {{Cityscapes Dataset}} for
  {{Semantic Urban Scene Understanding}},'' in \emph{{{IEEE Conf.}} on
  {{Computer Vision}} and {{Pattern Recognition}}}, 2016, pp. 3213--3223.

\bibitem{neuhold2017}
G.~Neuhold, T.~Ollmann, S.~R. Bulò, and P.~Kontschieder, ``{The Mapillary
  Vistas Dataset for Semantic Understanding of Street Scenes},'' in \emph{IEEE
  Int. Conf. Computer Vision}, 2017.

\bibitem{arnaba2018}
A.~Arnab, O.~Miksik, and P.~H.~S. Torr, ``On the {{Robustness}} of {{Semantic
  Segmentation Models}} to {{Adversarial Attacks}},'' in \emph{IEEE/CVF Conf.
  Computer Vision and Pattern Recognition}, 2018, pp. 888--897.

\bibitem{hendrycks2020}
D.~Hendrycks, N.~Mu, E.~D. Cubuk, B.~Zoph, J.~Gilmer, and B.~Lakshminarayanan,
  ``{{AugMix}}: {{A Simple Data Processing Method}} to {{Improve Robustness}}
  and {{Uncertainty}},'' in \emph{Int. Conf. Learning Representations}, 2020.

\bibitem{zhang2021a}
J.~Zhang, Y.~Zhang, and X.~Xu, ``{{ObjectAug}}: {{Object}}-level {{Data
  Augmentation}} for {{Semantic Image Segmentation}},'' in \emph{{IEEE/CVF}
  Conf. Computer Vision and Pattern Recognition}, 2021.

\bibitem{kamann2020}
C.~Kamann, B.~Güssefeld, R.~Hutmacher, J.~H. Metzen, and C.~Rother,
  ``Increasing the {{Robustness}} of {{Semantic Segmentation Models}} with
  {{Painting}}-by-{{Numbers}},'' in \emph{Euro. Conf. Computer Vision}, 2020,
  pp. 369--387.

\bibitem{geirhos2019}
R.~Geirhos, P.~Rubisch, C.~Michaelis, M.~Bethge, F.~Wichmann, and W.~Brendel,
  ``{{ImageNet}}-trained {{CNNs}} {Are Biased towards Texture; Increasing Shape
  Bias Improves Accuracy and Robustness},'' in \emph{Int. Conf. Learning
  Representations}, 2019.

\bibitem{jackson}
P.~T. Jackson, A.~Atapour-Abarghouei, S.~Bonner, T.~P. Breckon, and B.~Obara,
  ``Style {{Augmentation}}: {{Data Augmentation}} via {{Style
  Randomization}},'' in \emph{{{IEEE}}/{{CVF Conf.}} {{Computer Vision}} and
  {{Pattern Recognition}}}, 2019, p.~10.

\bibitem{kim2019}
B.~Kim, H.~Kim, K.~Kim, S.~Kim, and J.~Kim, ``Learning {{Not}} to {{Learn}}:
  {{Training Deep Neural Networks With Biased Data}},'' in \emph{{{IEEE}}/{{CVF
  Conf.}} {{Computer Vision}} and {{Pattern Recognition}}}, 2019, pp.
  9004--9012.

\bibitem{alvi2018}
M.~Alvi, A.~Zisserman, and C.~Nellaaker, ``Turning a {{Blind Eye}}: {{Explicit
  Removal}} of {{Biases}} and {{Variation}} from {{Deep Neural Network
  Embeddings}},'' in \emph{{{European Conf.}} {{Computer Vision}}
  {{Workshops}}}, 2018.

\bibitem{wang2019}
T.~Wang, J.~Zhao, M.~Yatskar, K.-W. Chang, and V.~Ordonez, ``Balanced
  {{Datasets Are Not Enough}}: {{Estimating}} and {{Mitigating Gender Bias}} in
  {{Deep Image Representations}},'' in \emph{{{IEEE}}/{{CVF Int. Conf.}}
  {{Computer Vision}}}.\hskip 1em plus 0.5em minus 0.4em\relax {IEEE}, 2019,
  pp. 5309--5318.

\bibitem{lecun1998}
Y.~Lecun, L.~Bottou, Y.~Bengio, and P.~Haffner, ``{Gradient-based Learning
  Applied to Document Recognition},'' in \emph{Proceedings of the IEEE},
  vol.~86, no.~11, 1998, pp. 2278--2324.

\bibitem{simonyan2014}
K.~Simonyan, A.~Vedaldi, and A.~Zisserman, ``Deep {{Inside Convolutional
  Networks}}: {{Visualising Image Classification Models}} and {{Saliency
  Maps}},'' in \emph{Int. Conf. Learning Representations Workshop}, 2014.

\bibitem{yosinski2015}
J.~Yosinski, J.~Clune, A.~Nguyen, T.~Fuchs, and H.~Lipson, ``Understanding
  {{Neural Networks Through Deep Visualization}},'' in \emph{Int. Conf. Machine
  Learning - Deep Learning Workshop}, 2015.

\bibitem{ros2016}
G.~Ros, L.~Sellart, J.~Materzynska, D.~Vazquez, and A.~M. Lopez, ``The
  {SYNTHIA} {Dataset}: {A Large Collection of Synthetic Images for Semantic
  Segmentation of Urban Scenes},'' in \emph{IEEE Conf. Computer Vision and
  Pattern Recognition}, 2016.

\bibitem{deng2009}
J.~Deng, W.~Dong, R.~Socher, L.-J. Li, K.~Li, and L.~Fei-Fei, ``{ImageNet}: {A
  Large-Scale Hierarchical Image Database},'' in \emph{IEEE Conf. Computer
  Vision and Pattern Recognition}, 2009, pp. 248--255.

\bibitem{simonyan2015}
K.~Simonyan and A.~Zisserman, ``{Very Deep Convolutional Networks for
  Large-Scale Image Recognition},'' in \emph{Int. Conf. Learning
  Representations}, 2015.

\bibitem{kingma2014adam}
D.~P. Kingma and J.~Ba, ``{Adam: A method for Stochastic Optimization},''
  \emph{arXiv preprint arXiv:1412.6980}, 2014.

\bibitem{hu2019}
X.~Hu, C.-W. Fu, L.~Zhu, and P.-A. Heng, ``Depth-attentional features for
  single-image rain removal,'' in \emph{{{IEEE}}/{{CVF Conf.}} {{Computer
  Vision}} and {{Pattern Recognition}}}, 2019.

\bibitem{sakaridis2018}
C.~Sakaridis, D.~Dai, and L.~Van~Gool, ``{Semantic Foggy Scene Understanding
  with Synthetic Data},'' in \emph{Euro. Conf. Computer Vision}, vol. 126,
  no.~9, 2018, pp. 973--992.

\bibitem{amini2019}
A.~Amini, A.~P. Soleimany, W.~Schwarting, S.~N. Bhatia, and D.~Rus,
  ``Uncovering and {{Mitigating Algorithmic Bias}} through {{Learned Latent
  Structure}},'' in \emph{{{AAAI}}/{{ACM Conf.}} {{AI}}, {{Ethics}}, and
  {{Society}}}, 2019, pp. 289--295.

\end{thebibliography}

\end{document}